\documentclass{llncs}
\usepackage{makeidx}  
\usepackage{epsf}
\begin{document}

\mainmatter              
\title{A Machine-Learning Approach to Estimating the Referential Properties of  Japanese Noun Phrases}
\titlerunning{ML Approach to Estimating Referential Properties of Japanese NPs}  
%
\author{Masaki Murata \and Kiyotaka Uchimoto \and Qing Ma \and Hitoshi Isahara}

\authorrunning{Masaki Murata et al.}   
%
\tocauthor{Masaki Murata(Communications Research Laboratory, MPT),
Kiyotaka Uchimoto(Communications Research Laboratory, MPT),
Qing Ma(Communications Research Laboratory, MPT),
Hitoshi Isahara(Communications Research Laboratory, MPT)}

\institute{Communications Research Laboratory, MPT,\\ 
2-2-2 Hikaridai, Seika-cho, Soraku-gun, Kyoto, 619-0289, Japan,\\
\email{\{murata,uchimoto,qma,isahara\}@crl.go.jp},\\ WWW home page:
\texttt{http://www-karc.crl.go.jp/ips/murata}}

\maketitle              

\def\None#1{}

\def\Small#1{}

\def\OK#1{}

\def\Out#1{}

\def\Noun#1{{\sf NP #1}}
\def\Verb#1{{\sf Verb #1}}

\begin{abstract}
The referential properties of noun phrases 
in the Japanese language, which has no articles, are useful for 
article generation in Japanese-English machine translation and 
for anaphora resolution in Japanese noun phrases. 
They are generally classified as 
generic noun phrases, definite noun phrases, and indefinite noun phrases. 
In the previous work, 
referential properties were estimated by 
developing rules that used clue words. 
If two or more rules were in conflict with each other, 
the category having the maximum total score given by the rules 
was selected as the desired category. 
The score given by each rule was established by hand, 
so the manpower cost was high. 
In this work, 
we automatically adjusted these scores 
by using a machine-learning method 
and succeeded in reducing the amount of manpower 
needed to adjust these scores. 
\end{abstract}
%

\section{Introduction}\label{intro}

To estimate the referential property of a noun phrase (NP) 
in the Japanese language, 
which does not have any articles, is one of the most difficult problems 
in natural language processing \cite{match}. 
The referential property of a noun phrase represents
how the noun phrase denotes the referent and 
is classified into the following three types: 

\begin{itemize}
\item[$\bullet$] 
  An indefinite NP --- 
  denotes
  an arbitrary member of the class of the noun phrase.

  (Ex.) \mbox{There are three \underline{dogs}.}\\

\item[$\bullet$]  
  A definite NP --- 
  denotes a contextually non-ambiguous member
  of the class of the noun phrase.

  (Ex.) \mbox{\underline{The dog} went away.}\\

\item[$\bullet$]  
  A generic NP --- denotes all members of the class of the noun phrase 
  or the class itself of the noun phrase. 

  (Ex.) \mbox{\underline{Dogs} are useful.}

  Note that ``dogs'' in this sentence denotes general dogs and 
  are classified generic. 
\end{itemize}

Estimating the referential properties 
of noun phrases in Japanese sentences is useful for 
(i) generating articles 
when translating Japanese nouns into English and (ii) 
estimating the referents of noun phrases. 

{
\begin{itemize}
\item[(i)] Article generation in machine translation

In the process of generating articles, 
when a noun phrase is estimated to be indefinite, 
it is given the indefinite article, ``a/an'', 
when it is singular, but is given no article 
when it is plural. 
When a noun phrase is estimated to be definite, 
it is given the definite article, ``the''. 
When a noun phrase is estimated to be generic, 
usage in terms of articles 
is generated by a method used for generic noun phrases 
(a generic noun phrase can be given a definite or an indefinite 
article or no article, and may also be in the plural form.)\footnote{
Bond et al. have actually used the referential properties of noun phrases 
in generating articles \cite{Bond_94}.} 
For example, {\it hon} (book) in the following sentence 
is a generic noun phrase. 
\begin{equation}
  \begin{minipage}[h]{10cm}
    \begin{tabular}[h]{@{ }c@{ }c@{ }c@{ }c@{ }}
      {\it hon-toiunowa} & {\it ningen-no} & {\it seichou-ni} & {\it kakasemasen} \\
      (book) & (human being) & (growth) & (be necessary)\\
      \multicolumn{4}{@{}l@{}}{(Books are necessary for the growth of a human being.)}\\
    \end{tabular}
  \end{minipage}
  \label{eqn:book_hito}
\end{equation}
So it can be translated as 
``a book,'' ``books,'' or ``the book'' in English. 
Note that in the following sentence, 
{\it hon} (book) is a definite noun phrase. 
\begin{equation}
  \begin{minipage}[h]{10cm}
    \begin{tabular}[h]{@{ }c@{ }c@{ }c@{ }c@{ }c@{ }}
      {\it kinou} & {\it boku-ga} & {\it kashita} & {\it hon-wa} & {\it yomimashitaka} \\
      (yesterday) & (I) & (lend) & (book) & (read)\\
      \multicolumn{5}{@{}l@{}}{(Did you read the book I lent you yesterday?)}\\
    \end{tabular}
  \end{minipage}
  \label{eqn:book_kasu}
\end{equation}
It can thus be translated as ``the book'' in English. 

\item[(ii)] Anaphora resolution

Only a definite noun phrase can refer to 
a previous noun phrase and this is very useful 
in anaphora resolution \cite{murata_coling98}. 
For example, in the following example, 
{\it hon} (book) in the second sentence is 
a generic noun phrase, 
so it cannot be referring to 
{\it hon} (book) in the first sentence. 

\begin{equation}
  \begin{minipage}[h]{10cm}
    \begin{tabular}[h]{@{ }c@{ }c@{ }c@{ }c@{ }c@{ }}
      {\it hon-wo} & {\it omiyage-ni} & {\it kaimashita.} \\
      (book) & (as a present) & (buy) \\
      \multicolumn{5}{@{}l@{}}{(I bought books as a present.)}\\
    \end{tabular}

    \vspace{0.2cm}

    \begin{tabular}[h]{@{ }c@{ }c@{ }c@{ }c@{ }}
      {\it hon-toiunowa} & {\it ningen-no} & {\it seichou-ni} & {\it kakasemasen} \\
      (book) & (human being) & (growth) & (be necessary)\\
      \multicolumn{4}{@{}l@{}}{(Books are necessary for the growth of a human being.)}\\
    \end{tabular}
  \end{minipage}
  \label{eqn:book_miyage}
\end{equation}

\end{itemize}}

As in the above explanation, 
the referential properties of noun phrases, 
i.e., generic, definite, and indefinite, are 
useful for article generation and anaphora resolution, 
and estimating them is a serious problem 
in natural language processing. 

In the conventional estimation of referential properties \cite{match}, 
heuristic rules (made by hand) 
using surface clue words are used for estimation. 
For example, 
in sentence (\ref{eqn:book_hito}) above 
the referential property is estimated to be generic 
by using a Japanese clue word {\it toiu-nowa}; 
in sentence (\ref{eqn:book_kasu}), 
the referential property of 
a noun phrase, {\it hon} (book), 
is estimated to be definite 
since the noun phrase is modified by 
an embedded sentence, {\it kinou boku ga kashita} (I lent you yesterday). 
In their work, 
86 heuristic rules were created. 
When plural rules conflicted and the rule used in estimation was 
ambiguous, 
the conflict was solved by using the scores given in the rules. 
These scores needed to be adjusted by hand 
in order to properly resolve conflicts. 

In the current work, to reduce human costs of previous research, 
we have used a machine-learning method to automatically 
adjust these rules. 
We selected 
the maximum entropy method (which is robust for sparse data problems) 
as the machine-learning method. 

\None{
\section{Categories of referential property}\label{sec:riron}
Referential property of a noun phrase here means
how the noun phrase denotes the subject. 
We classified noun phrases into the following three types
from the referential property.
{\scriptsize
\[\rm \mbox{\normalsize  NP}
 \left\{ \begin{array}{l}
     \rm \mbox{\normalsize  {\bf generic} NP}\\
         \mbox{\normalsize  {\bf non generic} NP}
            \left\{ \begin{array}{l}
              \rm \mbox{\normalsize  {\bf definite} NP}\\
                  \mbox{\normalsize  {\bf indefinite} NP}
            \end{array}
            \right.
\end{array}
\right.
\]
}
\paragraph{Generic noun phrase}
A noun phrase is classified as generic
when it denotes all members of the class of the noun phrase
or the class itself of the noun phrase.
For example, ``dogs'' in the following sentence is a generic noun phrase.
\begin{equation}
\mbox{\underline{Dogs} are useful.}
  \label{eqn:doguse}
\end{equation}
\paragraph{Definite Noun Phrase}
A noun phrase is classified as definite
when it denotes a contextually non-ambiguous member
of the class of the noun phrase.
For example, ``the dog'' in the following sentence is a definite noun phrase.
\begin{equation}
\mbox{\underline{The dog} went away.}
  \label{eqn:thedogaway}
\end{equation}
\paragraph{Indefinite Noun Phrase}
An indefinite noun phrase denotes
an arbitrary member of the class of the noun phrase.
For example, the following ``dogs'' is an indefinite noun phrase.
\begin{equation}
\mbox{There are three \underline{dogs}.}
  \label{eqn:threedogs}
\end{equation}}

\section{How to estimate referential property}\label{sec:decide}

\subsection{Method used in previous research}\label{sec:decide_pre}

The previous research gave each referential property 
two evaluation values, {\it possibility} and {\it value}, 
by using heuristic rules 
and estimated the referential property according to these values. 
Here, {\it possibility} is logically conjuncted and {\it value} is added. 
As a result, the referential property whose {\it possibility} is 1 
and whose {\it value} is maximum is estimated to be the desired one. 

\noindent Heuristic rules are given in the following forms:\\

\noindent
{
 ({\it condition for rule application}) \\
\hspace*{0.5cm}$\Longrightarrow$ \{ 
indefinite ({\it possibility, value}) \\
\hspace*{1.4cm}definite ({\it possibility, value}) \\
\hspace*{1.4cm}generic ({\it possibility, value}) \}\\
}

\noindent 
A surface expression, which contains a clue word for 
estimating the referential property, is written 
in {\it condition for rule application}. 
{\it Possibility} has a value of 1 when the 
categories indefinite, definite and generic 
are possible in the context checked by the condition.  
Otherwise, the {\it possibility} value is 0. 
{\it Value} means that 
a relative possibility value between 1 and 10 
(an integer) is given according to 
the plausibility of the condition that the {\it possibility} is 1.
A larger value means the plausibility is high.


Several rules can be applicable to a specific noun in a sentence.  
In this case, 
the possibility values for the individual categories are added, 
and the category for the noun is decided 
as the category with the highest sum of possibility values. 

86 rules were created. 
Some of the rules are given below.

{
\begin{enumerate}
\item[(1)] When a noun is modified by a referential pronoun, {\it kono} (this), {\it sono} (its), etc., 
then \,
\{\mbox{indefinite}  (0, 0)\footnote{ 
(a, b) means {\it possibility} (a) and {\it value} (b). 
} 
\mbox{definite}   (1, 2)  
\mbox{generic}  (0, 0)\}\\
\begin{tabular}{ccccc}
\underline{\it kono} & \underline{\it hon-wa} & {\it omoshiroi.}\\
(this) & (book) & (interesting)\\
\multicolumn{3}{l@{}}{(\underline{This book} is interesting.)}\\
\end{tabular}
\item[(2)] When a noun is accompanied by a particle, {\it wa}, and the 
predicate is in the past tense, 
then \,
\{\mbox{indefinite} (1, 0)
\mbox{definite}   (1, 3) 
\mbox{generic} (1, 1)\}\\
\begin{tabular}{ccccc}
\underline{\it inu-wa} & {\it mukou-he} & {\it itta.}\\
(dog) & (away there) & (went)\\
\multicolumn{3}{l@{}}{(\underline{The dog} went away.)}\\
\end{tabular}
\item[(3)] When a noun is accompanied by a particle, {\it wa}, and the 
predicate is in the present tense, 
then \,
\{\mbox{indefinite} (1, 0) 
\mbox{definite}   (1, 2) 
\mbox{generic} (1, 3)\}\\
\begin{tabular}{ccccc}
\underline{\it inu-wa} & {\it yakunitatsu} & {\it doubutsu}& {\it desu.}\\
(dog) & (useful) & (animal) & (is)\\
\multicolumn{3}{l@{}}{(\underline{Dogs} are useful animals.)}\\
\end{tabular}
\end{enumerate}}

\None{
There are many other expressions which provide some clues 
for the referential property of nouns, such as 
(i) the noun itself, {\it chikyuu} (the earth) [definite], \\
{\it uchuu} (the universe) [definite], etc., 
(ii) nouns modified by a numeral 
(Example: {\it kore-wa} (this) {\it issatsu-no} (one) 
\underline{\it hon-desu} (book) [indefinite]. 
(This is \underline{a book}.)), 
(iii) the same noun presented previously 
(Example: {\it kare-wa} (he) {\it jouyousha} (car) {\it to} (and) 
{\it torakku-wo} (truck) 
{\it ichidai-zutsu} (one each) {\it motteimasuga} (have), 
\underline{\it jouyousha}-{\it nidake} (car) [definite] 
{\it hoken-wo-kaketeimasu} (be insured). 
(He has a car and a truck, but only the car is insured.)), 
(iv) adverb phrases, {\it itsumo} (always)'', {\it nihon-dewa} (in Japan)'', etc. 
(Example: {\it nihon-dewa} (in Japan) 
\underline{\it shashou-wa} (conductor) [generic] 
{\it joukyaku} (passenger) {\it no} (of) 
{\it kippu-wo} (ticket) {\it shirabemasu} (check). 
(In Japan,\\
\underline{the conductor} checks the tickets of the passengers.)),
(v) verbs, {\it suki} (like), {\it tanoshimu} (enjoy), etc. 
(Example: {\it watashi-wa} (I) 
\underline{\it ringo-ga} (apple) [generic] 
{\it suki-desu} (like). 
(I like \underline{apples}.)). }

When there are no clues, ``indefinite'' is assigned 
as the default value.

Let us look at an example of a noun to which 
several rules apply, {\it kudamono} (fruit) 
as used in the following sentence.

\begin{equation}
  \begin{minipage}[h]{10cm}
    \begin{tabular}[t]{c@{ }c@{ }c@{ }c@{ }c@{ }c}
{\it wareware-ga} & {\it kinou} & {\it tsumitotta} & \underline{\it kudamono}-\underline{\it wa} & {\it aji-ga} & {\it iidesu.}\\
(we) & (yesterday) & (picked) & (fruit) &   (taste) & (be good)\\
\multicolumn{6}{l}{
(\underline{The fruit} that we picked yesterday tastes delicious.)}
\end{tabular}
  \end{minipage}
\label{eqn:kudamono}
\end{equation}

All the rules were applied and 
the condition only satisfied 
the following seven rules 
which were then used to determine 
the degree of the definiteness of the noun. 

{
\begin{itemize}
\item[(a)] When a noun is accompanied by {\it wa}, and the corresponding predicate 
is not in the past tense \\
({\it kudamono}-\underline{\it wa} {\it aji-ga} \underline{\it iidesu}), 
then \\
\{\mbox{indefinite}  (1, 0) 
\mbox{definite}  (1, 2)   
\mbox{generic}  (1, 3)\}

\item[(b)] When a noun is modified by an embedded sentence which is in the past tense ({\it tsumitotta}),\\
then \,\\
\{\mbox{indefinite}  (1,  0) 
\mbox{definite}    (1,  1) 
\mbox{generic}  (1,  0)\}

\item[(c)] When a noun is modified by an embedded sentence which has a 
definite noun accompanied by {\it wa} or {\it ga} ({\it wareware-ga}), 
 then \\
\{\mbox{indefinite}  (1,  0) 
\mbox{definite}  (1,  1) 
\mbox{generic}  (1,  0)\}

\item[(d)] When a noun is modified by an embedded sentence which has a definite noun accompanied by a particle ({\it wareware-ga}), 
 then \\
\{\mbox{indefinite}  (1,  0) 
\mbox{definite}  (1,  1) 
\mbox{generic}  (1,  0)\}

\item[(e)] When a noun is modified by a phrase which has a pronoun 
({\it wareware-ga}), 
then \\
\{\mbox{indefinite}  (1,  0) 
\mbox{definite}    (1,  1) 
\mbox{generic}  (1,  0)\}

\item[(f)] When a noun has an adjective as its predicate 
({\it kudamono-wa azi-ga} \underline{\it iidesu}), 
then \\
\{\mbox{indefinite}  (1,  0) 
\mbox{definite}    (1,  3) 
\mbox{generic}  (1,  4)\}

\item[(g)] When a noun is a common noun ({\it kudamono}),\\
then \\
\{\mbox{indefinite}  (1,  1) 
\mbox{definite}    (1,  0) 
\mbox{generic}  (1,  0)\}

\end{itemize}}

\None{
By applying all these rules, 
the final score of 
\{\mbox{indefinite}  (1,  1) 
\mbox{definite}    (1,  9) 
\mbox{generic}  (1,  7)\} 
were obtained for {\it kudamono} (fruit)
and ``definite'' was determined to be correct. 
}

By using the scores from these rules, 
total {\it possibilities} and 
{\it values} were calculated. 
The overall {\it possibilities} for all three categories were 1 because 
all three categories carried {\it possibilities} of 1 
for all of these rules. 
The total {\it values} for all three categories were 1, 9, and 7, respectively, 
because the results of aggregating the {\it values} of 
all of these rules were 
1 ($= 0 + 0 + 0 + 0 + 0 + 0 + 1$), 
9 ($= 2 + 1 + 1 + 1 + 1 + 3 + 0$), and 
7 ($= 3 + 0 + 0 + 0 + 0 + 4 + 0$), for the respective categories. 
A final score of 
\{\mbox{indefinite}  (1,  1) 
\mbox{definite}    (1,  9) 
\mbox{generic}  (1,  7)\} was obtained, 
and the system judged, correctly, that the noun here is 
``definite.''

Each noun, from left to right, in the sentence was 
estimated according to (a) - (g) above. 
This process allows the decision process to make use of 
referential properties that has already been determined 
(see (c) and (d), for example). 

In the method used in the previous work 
{\it possibility} and {\it value} 
had to be adjusted 
in order to estimate referential properties properly, 
and this required much work by human. 
Although gathering the clue words by hand 
might be effective, 
{\it possibility} and {\it value} 
can be adjusted by a certain machine-learning method. 
In this paper, 
we report on our use of the machine-learning method 
described in the next section. 
to verify this possibility, 

\subsection{The machine-learning method}\label{sec:decide_now}

A machine-learning method was applied to 
the estimation of referential properties. 
We used the maximum entropy method as a machine-learning method, 
which is robust against data sparseness, 
because it is difficult to make 
a large corpus tagged with referential properties. 
By defining a set of features in advance, 
the maximum entropy method estimates 
the conditional probability of each category 
in a certain situation of the features, 
and it is called {\it the maximum entropy method} 
since it maximizes entropy when estimating a probability. 
The process of maximizing entropy 
can make the probabilistic model uniform,  
and this effect is the reason that 
the maximum entropy method is robust against 
data sparseness. We used 
Ristad's system \cite{ristad97,ristad98} 
as the maximum entropy method. 
The three probabilities, 
generic, definite, or indefinite, are calculated 
from the output of Ristad's system. 
The category having the maximum probability 
is judged to be the desired one. 

To use the maximum entropy method, 
we must choose the features used in learning. 
We used the conditions of the 86 rules 
that had been used in the previous work. 
86 features are thus used in learning. 

If we use, for example, 
rules 1, 2, and 3 as described in Sec. \ref{sec:decide_pre}, 
only condition parts are detected and 
the following three features are obtained. 
{
\begin{enumerate}
\item Whether or not a noun is modified by a referential pronoun, 
{\it kono} (this), {\it sono} (its), etc.
\item Whether or not a noun is accompanied 
by a particle {\it wa}, and the 
predicate is in the past tense. 
\item Whether or not a noun is accompanied 
by a particle {\it wa}, and the 
predicate is in the present tense. 
\end{enumerate}}

Now, we use the last example from 
the previous section 
to explain how the referential property 
is estimated by the maximum entropy method. 

\begin{equation}
  \begin{minipage}[h]{10cm}
    \begin{tabular}[t]{c@{ }c@{ }c@{ }c@{ }c@{ }c}
{\it wareware-ga} & {\it kinou} & {\it tsumitotta} & \underline{\it kudamono}-\underline{\it wa} &{\it aji-ga} & {\it iidesu.}\\
(we) & (yesterday) & (picked) & (fruit) &  (taste) & (be good)\\
\multicolumn{6}{l}{
(\underline{The fruit} that we picked yesterday tastes delicious.)}
\end{tabular}
  \end{minipage}
\label{eqn:kudamono2}
\end{equation}

We again look at {\it kudamono} (fruit). 
The same seven rules are again applied. 
The value assigned to each of the referential properties 
for each rule indicates 
the conditional probability of that category being correct 
when only that rule is applied and 
they are calculated by the maximum entropy method. 
The values written here were obtained in 
our experiment of ``Machine-Learning 2'' 
described in Sec. \ref{sec:jikken}. 

{
\begin{itemize}
\item[(a)] When a noun is accompanied by {\it wa} and the corresponding predicate 
is not in the past tense, \\
({\it kudamono}-\underline{\it wa} {\it aji-ga} \underline{\it iidesu}), 
then \\
\{\mbox{indefinite}   0.31 \,
\mbox{definite}   0.29  \,
\mbox{generic}   0.40\}

\item[(b)] When a noun is modified by an embedded sentence which is in the past tense ({\it tsumitotta}),\\
then \,\\
\{\mbox{indefinite}  0.31 \,
\mbox{definite}     0.49 \, 
\mbox{generic}   0.19\}

\item[(c)] When a noun is modified by an embedded sentence which has a 
definite noun accompanied by {\it wa} or {\it ga} ({\it wareware-ga}), 
 then \\
\{\mbox{indefinite}  0.19  \,
\mbox{definite}   0.61 \,
\mbox{generic}  0.19\}

\item[(d)] When a noun is modified by an embedded sentence which has a definite noun accompanied by a particle ({\it wareware-ga}), 
 then \\
\{\mbox{indefinite} 0.01 \,
\mbox{definite}    0.80 \,
\mbox{generic}     0.18\}

\item[(e)] When a noun is modified by a phrase which has a pronoun 
({\it wareware-ga}), 
then \\
\{\mbox{indefinite}  0.20 \,
\mbox{definite}     0.44 \,
\mbox{generic}      0.37\}

\item[(f)] When a noun has an adjective as its predicate 
({\it kudamono-wa azi-ga} \underline{\it iidesu}), 
then \\
\{\mbox{indefinite}  0.13 \,
\mbox{definite}     0.80 \,
\mbox{generic}      0.07\}

\item[(g)] When a noun is a common noun ({\it kudamono}),\\
then \\
\{\mbox{indefinite}   0.72 \,
\mbox{definite}    0.15 \,
\mbox{generic}      0.14\}

\end{itemize}}

In the maximum entropy method 
the values assigned by the above rules 
are multiplied, 
the values in each category are normalized, 
and 
the category with the highest value is judged to be 
the desired one. 
In this case, 
we multiplied and normalized the values of all the rules 
and obtained the following results: \\
{
\mbox{\{\mbox{indefinite}  0.001, \,
\mbox{definite}    0.996, \,
\mbox{generic}  0.002\}}}\\
``Definite'' had the highest value and 
was thus judged to be the desired category. 

\section{Experiment and Discussion}\label{sec:jikken}

Morphological and syntactic information are used 
as features in estimating referential properties. 
Before estimating the referential property, 
morphology and syntax were analyzed \cite{JUMAN3.5_e,csan2_ieice}. 
We used the same learning set and test set 
as had been used in previous work.\footnote{The learning set: 
``Usage of the English Articles''\cite{kanshi_eng}, 
a folktale ``The Old Man with a Wen'' \cite{kobu_eng},
and an essay {\it tensei jingo}. 
The test set: a folktale {\it turu no ongaeshi} \cite{kobu_eng},
an essay {\it tensei jingo}, 
 ``Pacific Asia in the Post-Cold-War World''
(A Quarterly Publication of The International House of Japan Vol. 12, 
No. 2 Spring 1992).} 
The 86 rules had been made by examining 
the learning set by hand. 
In the previous work, 
the values written in the 86 rules had been adjusted 
by checking the accuracy rates in the learning set. 

Firstly, we carried out the following two experiments. 
{
\begin{itemize}
\item 
  Manual Adjustment --- 
  The estimation was made by using the method described in Sec \ref{sec:decide_pre}. 
  (This result is identical to the result of the previous work.)

\item 
  Machine-Learning 1 ---
  The estimation was made by using the method described in Sec \ref{sec:decide_now}. 

\end{itemize}}
The results are listed in Tables \ref{tab:kanshi_d} to \ref{tab:turu_m1}. 
``Other'' in the tables indicates that 
the referential property is ambiguous; 
such cases can be neglected here since 
they were few.

\begin{table}[p]

\caption{Manual Adjustment (learning set)}\label{tab:kanshi_d}

\begin{center}

{

\begin{tabular}[c]{|l@{ }|@{ }r@{ }|r@{ }|r@{ }|r@{ }|r@{ }|} \hline
\multicolumn{1}{|c|}{}  &  indef  & def &  gen &  other & total \\\hline
\multicolumn{6}{|@{}c@{}|}{Usage of the Articles (140 sentences, 380 nouns)} \\ \hline 
correct &      96  &     184  &      58  &       1  &     339  \\
incorrect  &       4  &      28  &       8  &       1  &      41  \\
\hline 
\% of correct&    96.0  &    86.8  &    87.9  &    50.0  &    89.2  \\\hline
\multicolumn{6}{|@{}c@{}|}{The Old Man with a Wen (104 sentences, 267 nouns)} \\ \hline 
correct  &      73  &     140 &       6  &       1  &    222  \\
incorrect &      14  &      27  &       4  &       0  &      45    \\
\hline
\% of correct&   83.9  &   84.0  &   60.0  &  100.0  &   83.2  \\\hline
\multicolumn{6}{|@{}c@{}|}{an essay {\it tensei jingo} (23 sentences, 98 nouns)} \\ \hline 
correct  &      25  &     35 &      16  &       0  &    76  \\
incorrect  &       5  &      14  &       3  &       0  &      22 \\
\hline
\% of correct&    83.3  &    71.4  &    84.2  &    -----  &    77.6   \\\hline
average &    &    &    &    &     \\
\, \% of appearance &  29.1  &  57.7   &  12.8   &  0.4  &  100.0   \\
\, \% of correct &  89.4  &  84.0   &  84.2   &  66.7   &  85.5  \\\hline
\end{tabular}
}
\end{center}
\end{table}

\begin{table}[p]

  \caption{Manual Adjustment (test set)}\label{tab:turu_d}

\begin{center}

{

\begin{tabular}[c]{|l@{ }|@{ }r@{ }|r@{ }|r@{ }|r@{ }|r@{ }|} \hline
\multicolumn{1}{|c|}{}  &  indef  & def &  gen &  other & total \\\hline
\multicolumn{6}{|c|}{a folktale {\it Turu} (263 sentences, 699 nouns)} \\ \hline 
correct  &     109  &    363  &      13  &      10  &   495   \\
incorrect  &      38  &     160  &       6  &       0  &     204 \\
\hline
\% of correct  &    74.2  &    69.4  &    68.4  &   100.0  &    70.8 \\\hline
\multicolumn{6}{|@{}c@{}|}{an essay {\it tensei jingo} (75 sentences, 283 nouns)} \\ \hline 
correct  &      75  &    81  &      16  &       0  &   172  \\
incorrect  &      41  &      60  &      10  &       0  &     111  \\
\hline
\% of correct  &    64.7  &    57.5  &    61.5  &    ----- &    60.8   \\\hline
\multicolumn{6}{|c|}{Pacific Asia (22 sentences, 192 nouns)} \\\hline 
correct   &      21  &    108  &      11  &       2  &  142  \\
incorrect &      17  &      31  &       2  &       0  &      50  \\
\hline
\% of correct   &    55.3  &    77.7  &    84.6  &   100.0  &    74.0   \\\hline
average &    &    &    &    &     \\
\, \% of appearance &  25.6  &  68.4  &  4.9   &  1.0   &  100.0 \\
\, \% of correct  &  68.1   &  68.7 &  69.0 &  100.0  &  68.9  \\\hline
\end{tabular}
}
\end{center}
\end{table}

\begin{table}[p]

\caption{Machine Learning 1 (learning set)}\label{tab:kanshi_m1}

\begin{center}

{

\begin{tabular}[c]{|l@{ }|@{ }r@{ }|r@{ }|r@{ }|r@{ }|r@{ }|} \hline
\multicolumn{1}{|c|}{}  &  indef  & def &  gen &  other & total \\\hline
\multicolumn{6}{|@{}c@{}|}{Usage of the Articles (140 sentences, 380 nouns)} \\ \hline 
correct  &      95  &     199  &      32  &       0  &     326   \\
incorrect &       5  &      13  &      34  &       2  &      54   \\\hline
\% of correct &   95.0  &   93.9  &   48.5  &    0.0  &   85.8   \\\hline
\multicolumn{6}{|@{}c@{}|}{The Old Man with a Wen (104 sentences, 267 nouns)} \\ \hline 
correct   &      71  &     151  &       1  &       0  &     223   \\
incorrect  &      16  &      18  &       9  &       1  &      44   \\\hline
\% of correct  &   81.6  &   89.4  &   10.0  &    0.0  &   83.5   \\\hline
\multicolumn{6}{|@{}c@{}|}{an essay {\it tensei jingo} (23 sentences, 98 nouns)} \\ \hline 
correct  &      21  &      46  &       5  &       0  &      72   \\
incorrect &       9  &       3  &      14  &       0  &      26   \\\hline
\% of correct  &   70.0  &   93.9  &   26.3  &     ---  &   73.5   \\\hline
average &    &    &    &    &     \\
\, \% of appearance  &  29.1  &  57.7   &  12.8   &  0.4  &  100.0   \\
\, \% of correct  &   86.2  &   92.1  &   40.0  &    0.0  &   83.4   \\\hline
\end{tabular}
}
\end{center}
\end{table}

\begin{table}[p]

\caption{Machine Learning 1 (test set)}\label{tab:turu_m1}

\begin{center}

{

\begin{tabular}[c]{|l@{ }|@{ }r@{ }|r@{ }|r@{ }|r@{ }|r@{ }|} \hline
\multicolumn{1}{|c|}{}  &  indef  & def &  gen &  other & total \\\hline
\multicolumn{6}{|c|}{a folktale {\it Turu} (263 sentences, 699 nouns)} \\ \hline 
correct   &     104  &     408  &       0  &       0  &     512   \\
incorrect  &      43  &     115  &      19  &      10  &     187   \\\hline
\% of correct &   70.8  &   78.0  &    0.0  &    0.0  &   73.3   \\\hline
\multicolumn{6}{|@{}c@{}|}{an essay {\it tensei jingo} (75 sentences, 283 nouns)} \\ \hline 
correct   &      72  &     108  &       2  &       0  &     182   \\
incorrect  &      44  &      33  &      24  &       0  &     101   \\\hline
\% of correct  &   62.1  &   76.6  &    7.7  &     ---  &   64.3   \\\hline
\multicolumn{6}{|c|}{Pacific Asia (22 sentences, 192 nouns)} \\\hline 
correct   &      21  &     130  &       1  &       0  &     152   \\
incorrect  &      17  &       9  &      12  &       2  &      40   \\\hline
\% of correct  &   55.3  &   93.5  &    7.7  &    0.0  &   79.2   \\\hline
average &    &    &    &    &     \\
\, \% of appearance &  25.6  &  68.4  &  4.9    &  1.0   &  100.0 \\
\, \% of correct  &   65.5  &   80.5  &    5.2  &    0.00  &   72.1   \\\hline
\end{tabular}
}
\end{center}
\end{table}

The accuracy rate obtained by Manual Adjustment (Table \ref{tab:turu_d}) 
in the test set was 68.9\% and 
that of Machine-Learning 1 (Table \ref{tab:turu_m1}) was 72.1\%. 
The machine-learning method was thus more accurate 
than Manual Adjustment method. 
But, as can be seen in Tables \ref{tab:turu_d} and \ref{tab:turu_m1}, 
all the categories of Manual Adjustment are about 70\% 
and the method did not have a bad-accuracy category. 
However, the result from Machine-Learning 1 of the ``generic'' category 
was low (5.2\%). 
We cannot therefore conclude that 
Machine-Learning 1 reliably gives good results\footnote{Here, we assume that 
not producing a bad-accuracy category was more important than 
having the highest total accuracy rate. 
We then constructed Method 2 as described in the following passages. 
However, if we assume that 
having the highest total accuracy rate is more important than 
not producing a bad-accuracy category, 
Method 1 is, in fact, better.}. 

\begin{table}[p]

\caption{Machine Learning 2 (learning set)}\label{tab:kanshi_m2}

\begin{center}

{

\begin{tabular}[c]{|l@{ }|@{ }r@{ }|r@{ }|r@{ }|r@{ }|r@{ }|} \hline
\multicolumn{1}{|c|}{}  &  indef  & def &  gen &  other & total \\\hline
\multicolumn{6}{|@{}c@{}|}{Usage of the Articles (140 sentences, 380 nouns)} \\ \hline 
correct   &      97  &     188  &      57  &       0  &     342   \\
incorrect  &       3  &      24  &       9  &       2  &      38   \\\hline
\% of correct  &   97.0  &   88.7  &   86.4  &    0.0  &   90.0   \\\hline
\multicolumn{6}{|@{}c@{}|}{The Old Man with a Wen (104 sentences, 267 nouns)} \\ \hline 
correct   &      80  &     137  &       6  &       0  &     223   \\
incorrect  &       7  &      32  &       4  &       1  &      44   \\\hline
\% of correct   &   92.0  &   81.1  &   60.0  &    0.0  &   83.5   \\\hline
\multicolumn{6}{|@{}c@{}|}{an essay {\it tensei jingo} (23 sentences, 98 nouns)} \\ \hline 
correct  &      26  &      40  &      17  &       0  &      83   \\
incorrect  &       4  &       9  &       2  &       0  &      15   \\\hline
\% of correct   &   86.7  &   81.6  &   89.5  &     ---  &   84.7   \\\hline
average &    &    &    &    &     \\
\, \% of appearance &  29.1  &  57.7   &  12.8   &  0.4  &  100.0   \\
\, \% of correct &   93.6  &   84.9  &   84.2  &    0.0  &   87.0   \\\hline
\end{tabular}
}
\end{center}
\end{table}

\begin{table}[p]

\caption{Machine Learning 2 (test set)}\label{tab:turu_m2}

\begin{center}

{
\begin{tabular}[c]{|l@{ }|@{ }r@{ }|r@{ }|r@{ }|r@{ }|r@{ }|} \hline
\multicolumn{1}{|c|}{}  &  indef  & def &  gen &  other & total \\\hline
\multicolumn{6}{|c|}{a folktale {\it turu} (263 sentences, 699 nouns)} \\ \hline 
correct   &     112  &     360  &      13  &       0  &     485   \\
incorrect  &      35  &     163  &       6  &      10  &     214   \\\hline
\% of correct  &   76.2  &   68.8  &   68.4  &    0.0  &   69.4   \\\hline
\multicolumn{6}{|@{}c@{}|}{an essay {\it tensei jingo} (75 sentences, 283 nouns)} \\ \hline 
correct   &      79  &      88  &      14  &       0  &     181   \\
incorrect  &      37  &      53  &      12  &       0  &     102   \\\hline
\% of correct  &   68.1  &   62.4  &   53.9  &     ---  &   64.0   \\\hline
\multicolumn{6}{|c|}{Pacific Asia (22 sentences, 192 nouns)} \\\hline 
correct  &      25  &     110  &      10  &       0  &     145   \\
incorrect  &      13  &      29  &       3  &       2  &      47   \\\hline
\% of correct  &   65.8  &   79.1  &   76.9  &    0.0  &   75.5   \\\hline
average &    &    &    &    &     \\
\, \% of appearance &  25.6  &  68.4 &  4.9    &  1.0   &  100.0 \\
\, \% of correct  &   71.8  &   69.5  &   63.8  &    0.0  &   69.1   \\\hline
\end{tabular}
}
\end{center}
\end{table}

We felt that the reason for this 
low accuracy in estimating the ``generic'' category is that 
the frequency of ``generic'' terms is low and 
machine learning is biased toward ``definite'' terms, 
which have higher frequency than ``generic'' terms. 
We therefore carried out the following further experiments. 
{
\begin{itemize}
\item 
  Machine-Learning 2 --- 
  When machine learning is performed 
  by using the maximum entropy method, 
  the number of events in each category of the learning set is 
  multiplied by the inverse of its occurrence. 
  For example, in this paper, 
  we multiplied 4, 2, and 9 by the frequencies 
  of ``indefinite,'' ``definite,'' 
  and ``generic.'' 
\end{itemize}}

In other words, 
since generic noun phrases only made up $2/9$ of definite noun phrases, 
we changed the frequencies of data 
as if definite noun phrases had occurred twice as often as 
in the actual data 
and generic noun phrases had occurred nine times more often 
than in the actual data. 
We found that 
this change made the frequencies of the three referential properties 
uniform and did not bias the analysis towards 
definite noun phrases. 
The change produced the following result. 
Machine-Learning 1 applies a general method of 
learning and learns data in order to maximize the following equation. 
\begin{equation}
  \label{eq:ml1}
  \mbox{evaluation function = [\% of correct in overall the data]}
\end{equation}
On the other hand, 
Machine-Learning 2 uses the frequencies of the referential properties 
and learns data in order to maximize the following equation. 
\begin{equation}
  \label{eq:ml2}
  \begin{minipage}[h]{6.5cm}
\begin{tabular}[h]{l@{}l}
\multicolumn{2}{l}{evaluation function} \\
\,  = &the average of \\
 &[\% of correct in ``indefinite''], \\
 &[\% of correct in ``definite''] and \\
 &[\% of correct in ``generic''] \\
\end{tabular}
  \end{minipage}
\end{equation}

The results for Machine-Learning 2 are listed in 
Tables \ref{tab:kanshi_m2} and \ref{tab:turu_m2}.

The accuracy rate of Machine-Learning 2 on the test set 
was 69.1\%. 
This is nearly equal to the 68.9\% 
obtained by using the Manual Adjustment. 
It was found that 
the accuracy rates for all three referential properties 
were about 70\%. 
Since even for the worst category, ``generic,'' 63.8\% was 
achieved, it is clear that Machine-Learning 2 
was able to quite precisely estimate the referential properties. 

\Small{
These experiments taught us the following. 
\begin{itemize}
\item 
  In the estimation of referential properties, 
  manual adjustment of the values of rules for solving conflicts 
  is not necessary when using the machine learning method. 
\item 
  In the estimation of referential properties, 
  to make the frequencies of categories uniformly 
  makes the accuracy rates uniformly. 
\end{itemize}}

As the above results show, 
we found that 
manual adjustment for estimating referential properties 
was not necessary and 
therefore human costs were decreased. 

\None{
But, this paper only deals with solving rule conflicts, 
and determining which expression is effective as 
a clue word, is still left to human beings. 
Although we are going to develop a method that 
automatically detects effective clue words, 
we also think that 
the extraction of effective clue words is difficult, 
and it has to be done by human beings. }

We also examined the values of the rules as given by 
the maximum entropy method used in Machine-Learning 2. 
We examine some of the rules as listed below. 
In each of the rules in the list, (i) the condition parts, 
(ii) values assigned by hand, 
and (iii) values assigned by Machine learning 2 are included. 
{
\begin{enumerate}
\item 
  Rules for indefinite noun phrases 

{
  (a) When a noun is accompanied by a particle, {\it ga} (new-topic marker), then\\
    \mbox{\{\mbox{indefinite} (1, 2) 
      \mbox{definite}   (1, 1) 
      \mbox{generic} (1, 0)\}}\\
    \mbox{\{\mbox{indefinite}   0.62, \,
      \mbox{definite}    0.21, \,
      \mbox{generic}   0.17\}}}

    \vspace{0.1cm}

    In general, a noun accompanied by a particle, {\it ga}, 
    one function of which is to indicate a new topic, 
    roughly tends to be indefinite. 
    The highest value is thus assigned to ``indefinite'' 
    when manual adjustment is used. 
    The value of ``indefinite'' as 
    determined by Machine-Learning 2 is also the highest. 
    In manual adjustment, 
    the possibilities of all categories are set to 1, 
    so any of the three categories can be the answer.  
    In Machine-Learning 2, 
    since the value of ``indefinite'' is not terribly high, 
    (not, e.g., 0.99), 
    any of the three categories can also be the answer. 

    \vspace{0.1cm}

{
    (b) When a noun is modified by an adjective {\it aru} (a certain $\sim$), then\\
    \mbox{\{\mbox{indefinite} (1, 2) \,
      \mbox{definite}   (0, 0) \,
      \mbox{generic} (0, 0)\}}\\
    \mbox{\{\mbox{indefinite} \ 0.99, 
      \mbox{definite}   {0.0001}, 
      \mbox{generic}  {0.0001}\}}}

    \vspace{0.1cm}

    Generally, a noun modified by {\it aru} (a certain $\sim$) is indefinite. 
    The values for {\it possibility} of the other categories 
    are thus set to 0 in manual adjustment. 
    The values used in Machine-Learning 2 are 
    0.0001, which is also a very small value. 
    The value of ``indefinite'' is extremely high in 
    Machine-Learning 2. 
    We find that Machine-Learning 2 can judge that 
    the category of a noun modified by {\it aru} is 
    almost certainly indefinite. 

\None{
    \vspace{0.2cm}
    (c)When a noun is a common noun, then\\
    \mbox{\{\mbox{indefinite} (1, 1) \,
      \mbox{definite}   (1, 0) \,
      \mbox{generic} (1, 0)\}}\\
    \mbox{\{\mbox{indefinite} \  0.72, \,
      \mbox{definite} \   0.15, \,
      \mbox{generic} \ 0.14\}}
    \vspace{0.1cm}
    This rule is a default rule which can 
    estimate that a noun is indefinite 
    when there are no clue words 
    and other rules are not applied. 
    The value of ``indefinite'' in Machine-Learning 2 
    is comparatively high, and 
    it effectively plays the role of a default rule appropriately. }

\item 
  Rules for definite noun phrases 

{
  (a)When a noun is a pronoun, then\\
    \mbox{\{\mbox{indefinite} (0, 0) \,
      \mbox{definite}   (1, 2) \,
      \mbox{generic} (0, 0)\}}\\
    \mbox{\{\mbox{indefinite}   0.005, \,
      \mbox{definite} \  0.99, \,
      \mbox{generic}   0.005\}}}

  When a noun is a pronoun, it is always definite, 
  so the values of {\it possibility} of the other categories 
  are manually set to 0. 
  The values assigned to the other categories 
  by Machine-Learning 2 are also very small. 

    \vspace{0.1cm}

{
  (b) When a noun is modified by an embedded sentence which has a 
  definite noun accompanied by {\it wa} or {\it ga} (nominative-case particle), then\\
  \mbox{\{\mbox{indefinite} (1, 0) \,
      \mbox{definite}   (1, 1) \,
      \mbox{generic} (1, 0)\}}\\
    \mbox{\{\mbox{indefinite} \  0.19, \,
      \mbox{definite}  \   0.61, \,
      \mbox{generic} \  0.19\}}}

  Although such a noun is not always definite, 
  it is likely to be definite. 
  The value of ``definite'' in Machine-Learning 2 
  is the highest of the three, but is not extremely high. 

\None{
    \vspace{0.1cm}
    (c) When the same noun appears in the previous five sentences 
    and is definite, then\\
    \mbox{\{\mbox{indefinite} (1, 1) \,
      \mbox{definite}   (1, 3) \,
      \mbox{generic} (1, 0)\}}\\
    \mbox{\{\mbox{indefinite}  \ 0.07, \,
      \mbox{definite}   \  0.88, \,
      \mbox{generic} \  0.05\}}
    \vspace{0.1cm}
    A noun often refers to 
    the same noun appearing ahead which is apt to be definite. 
    The value of ``definite'' in Machine-Learning 2 is high. }

\item 
  Rules for generic noun phrases 

{
  (a) When a noun is followed by a particle {\it wa}, 
  which does not have a modifier, then\\
  \mbox{\{\mbox{indefinite} (1, 0) \,
    \mbox{definite}   (1, 1) \,
    \mbox{generic} (1, 1)\}}\\
  \mbox{\{\mbox{indefinite} \ 0.03, \,
    \mbox{definite} \  0.26, \,
    \mbox{generic} \ 0.71\}}}

    \vspace{0.1cm}

    The particle {\it wa}, which is a topic marker,
    is an expression which 
    is likely to accompany either a definite noun phrase or 
    a generic noun phrase. 
    The {\it possibility value}s  of the two categories 
    are both set to 1 by Manual Adjustment. 
    Machine-Learning 2 found that 
    the values assigned to ``definite'' and 
    ``generic'' are higher than that assigned to 
    ``indefinite,'' 
    but ``generic'' is assigned a higher value than ``definite.'' 
    This is because there are many rules for estimating 
    that a noun phrase is ``definite,'' 
    so a word can be estimated as ``generic'' 
    if no other clue words appear. 

    \vspace{0.1cm}

{
  (b) When a noun is followed by a particle {\it wa} 
  and it modifies an adjective, then \\
  \mbox{\{\mbox{indefinite}  (1, 0) \,
    \mbox{definite}    (1, 3) \,
    \mbox{generic}  (1, 4)\}}\\
  \mbox{\{\mbox{indefinite} \ 0.13, \,
    \mbox{definite}  \  0.80, \,
    \mbox{generic} \  0.07\}}}

    \vspace{0.1cm}

    Although ``generic'' is assigned the highest value by Manual Adjustment, 
    ``definite'' is assigned the highest value by Machine-Learning 2. 
    This is because of (i) 
    a wrong estimate by Machine-Learning 2 due to a small learning set, 
    (ii) the influence of other rules, such as the previous rule, 
    or (iii) incorrect manual adjustment in the earlier work. 
    If the actual reason is (i), 
    then making the learning set larger should improve the results.

\None{
  (c) When a noun is followed by a particle {\it towa} or {\it toiunowa}, then \\
  \mbox{\{\mbox{indefinite} (0, 0) \,
      \mbox{definite}   (1, 0) \,
      \mbox{generic} (1, 2)\}}\\
    \mbox{\{\mbox{indefinite} \  0.05, \,
      \mbox{definite}  \   0.05, \,
      \mbox{generic}  \  0.90\}}
    \vspace{0.1cm}
    A noun followed by a particle {\it towa} or {\it toiunowa} 
    is often a generic noun phrase. 
    The values in Machine-Learning 2 have the tendency. 
    \vspace{0.2cm}
}
    
\None{
    (c) When a noun modifies a verb modified by a adverb such as 
    ``{\it itsumo} (always),'' ``{\it ippan-ni} (generally),'' 
    ``{\it dentouteki-ni} (traditionally)'' and 
    ``{\it mukashi-wa} (long ago),'' then\\
    \mbox{\{\mbox{indefinite} (0, 0) \,
      \mbox{definite}   (1, 1) \,
      \mbox{generic} (1, 2)\}}\\
    \mbox{\{\mbox{indefinite} \  0.02, \,
      \mbox{definite}  \   0.09, \,
      \mbox{generic} \  0.89\}}
    \vspace{0.1cm}
    In this case, 
    a noun is apt to be generic 
    because of generic adverbs such as ``always.'' 
    The value of ``generic'' in Machine-Learning 2 is also high. }

\end{enumerate}}

As stated above, 
the values assigned by Machine-Learning 2 tended to be 
similar to those obtained by hand, and 
they demonstrated some degree of linguistic intuition. 

\section{Conclusions}\label{sec:end}
We have succeeded in creating a system for automatically giving rules 
values for solving conflicts 
when estimating the referential properties of noun phrases. 
We have thus shown 
that the cost in terms of human time of manually adjusting values 
is not unavoidable. 
We also found that, in machine learning, 
making the frequencies of the categories uniform 
can help to make their accuracy rates more uniform. 
Finally, we examined the values produced for rules 
by applying the Machine-Learning method, and 
confirmed that they were consistent with 
linguistic intuition. 

\bibliographystyle{plain}
\bibliography{mysubmit}

\end{document}